\documentclass[journal]{IEEEtran}
\usepackage{graphicx}
\usepackage{cite}
\usepackage{picinpar}
\usepackage{amsmath}
\usepackage{url}
\usepackage{flushend}
\usepackage[latin1]{inputenc}
\usepackage{colortbl}
\usepackage{soul}
\usepackage{multirow}
\usepackage{pifont}
\usepackage{color}
\usepackage{alltt}
\usepackage[hidelinks]{hyperref}
\usepackage{enumerate}
\usepackage{siunitx}
\usepackage{breakurl}
\usepackage{epstopdf}
\usepackage{pbox}
\usepackage{subcaption}
\usepackage{amsfonts}
\usepackage{bm}
\usepackage{balance}
\usepackage{amssymb}

\begin{document}

\title{Underwater Visual-Inertial-Acoustic-Depth SLAM with DVL Preintegration for Degraded Environments}

\author{
	\vskip 1em
	
	Shuoshuo Ding, Tiedong Zhang, Dapeng Jiang, and Ming Lei

	\thanks{
	
		(Corresponding author: Tiedong Zhang.).
		
		Shuoshuo Ding, Tiedong Zhang and Dapeng Jiang are with School of Ocean Engineering and Technology \& Southern Marine science and Engineering Guangdong Laboratory (Zhuhai), Sun Yat-sen University, Zhuhai 519082, China, with Guangdong Provincial Key Laboratory of Information Technology for Deep Water Acoustics, Zhuhai 519082, China, and also with Key Laboratory of Comprehensive Observation of Polar Environment (Sun Yat-sen University), Ministry of Education, Zhuhai 519082, China (e-mail: dingshsh5@mail2.sysu.edu.cn, zhangtd5@mail.sysu.edu.cn, jiangdp5@mail.sysu.edu.cn).
		
		Ming Lei is with the School of Marine Science and Technology Domain, Beijing Institute of Technology (Zhuhai), Zhuhai 519088, China (e-mail: 6220250056@bitzh.edu.cn).
	}
}

\maketitle
	
\begin{abstract}
Visual degradation caused by limited visibility, insufficient lighting, and feature scarcity in underwater environments presents significant challenges to visual-inertial simultaneous localization and mapping (SLAM) systems. To address these challenges, this paper proposes a graph-based visual-inertial-acoustic-depth SLAM system that integrates a stereo camera, an inertial measurement unit (IMU), the Doppler velocity log (DVL), and a pressure sensor. The key innovation lies in the tight integration of four distinct sensor modalities to ensure reliable operation, even under degraded visual conditions. To mitigate DVL drift and improve measurement efficiency, we propose a novel velocity-bias-based DVL preintegration strategy. At the frontend, hybrid tracking strategies and acoustic-inertial-depth joint optimization enhance system stability. Additionally, multi-source hybrid residuals are incorporated into a graph optimization framework. Extensive quantitative and qualitative analyses of the proposed system are conducted in both simulated and real-world underwater scenarios. The results demonstrate that our approach outperforms current state-of-the-art stereo visual-inertial SLAM systems in both stability and localization accuracy, exhibiting exceptional robustness, particularly in visually challenging environments.
\end{abstract}

\begin{IEEEkeywords}
Underwater localization, simultaneous localization and mapping (SLAM), sensor fusion, DVL preintegration. 
\end{IEEEkeywords}

\markboth{}%
{}

\definecolor{limegreen}{rgb}{0.2, 0.8, 0.2}
\definecolor{forestgreen}{rgb}{0.13, 0.55, 0.13}
\definecolor{greenhtml}{rgb}{0.0, 0.5, 0.0}

\section{Introduction}

\IEEEPARstart{H}{uman} activities in the fields of ocean engineering and marine science are increasing steadily, encompassing scientific expeditions to study underwater hydrothermal vents and archaeological sites, inspections and maintenance of subsea pipelines and reservoirs, and salvage operations for wrecked aircraft and vessels. The unique nature of these operational areas renders them either inaccessible or extremely hazardous for humans. It is evident that unmanned underwater vehicles (UUVs) that offer mobility and autonomy represent a dependable solution. Acquiring precise vehicle pose estimation during UUV deployment and operation is imperative for successfully executing missions in complex and unexplored underwater environments \cite{bai2025sio}, \cite{ma2020efficient}, \cite{xie2025neurss}, \cite{rahman2022svin2}.

In recent years, simultaneous localization and mapping (SLAM) technology has enabled autonomous navigation for mobile robots and has become a research hotspot in the field of unmanned underwater vehicles (UUVs). Underwater SLAM based on vision has also received extensive research attention \cite{ding2024robust}. However, the harsh underwater environment often presents significant challenges to pure visual SLAM systems, including weak textures, low light conditions, and high dynamics. The integration of an inertial measurement unit (IMU), known as visual-inertial SLAM (VI-SLAM), has been demonstrated to mitigate the transient and noise-affected visual inputs from optical cameras \cite{ou2024hybrid}.

A unified underwater visual-inertial odometry method combining both direct and feature-based approaches was proposed \cite{miao2021univio}. This strategy simultaneously optimizes reprojection error, photometric error, and IMU error to achieve robust and high-precision odometry results. To address the challenge of poor underwater visibility, Wang et al. \cite{wang2023robust} proposed a method that combines the simultaneous extraction of point and diagonal features from images with inertial measurements to handle low-visibility underwater scenarios. To fully leverage pressure measurement information, a reverse IMU preintegration method was proposed \cite{hu2022tightly}. Pressure factors were derived using either forward or backward preintegration, depending on the time offset between pressure measurements and adjacent keyframes. To mitigate the impact of low-quality images on the system, image factors were integrated into the VI-odometry (VIO) based on relative depth measurements, with these factors dynamically weighted \cite{ding2023rd}.

\begin{figure}[t]
	\centering
	\includegraphics[width=\linewidth]{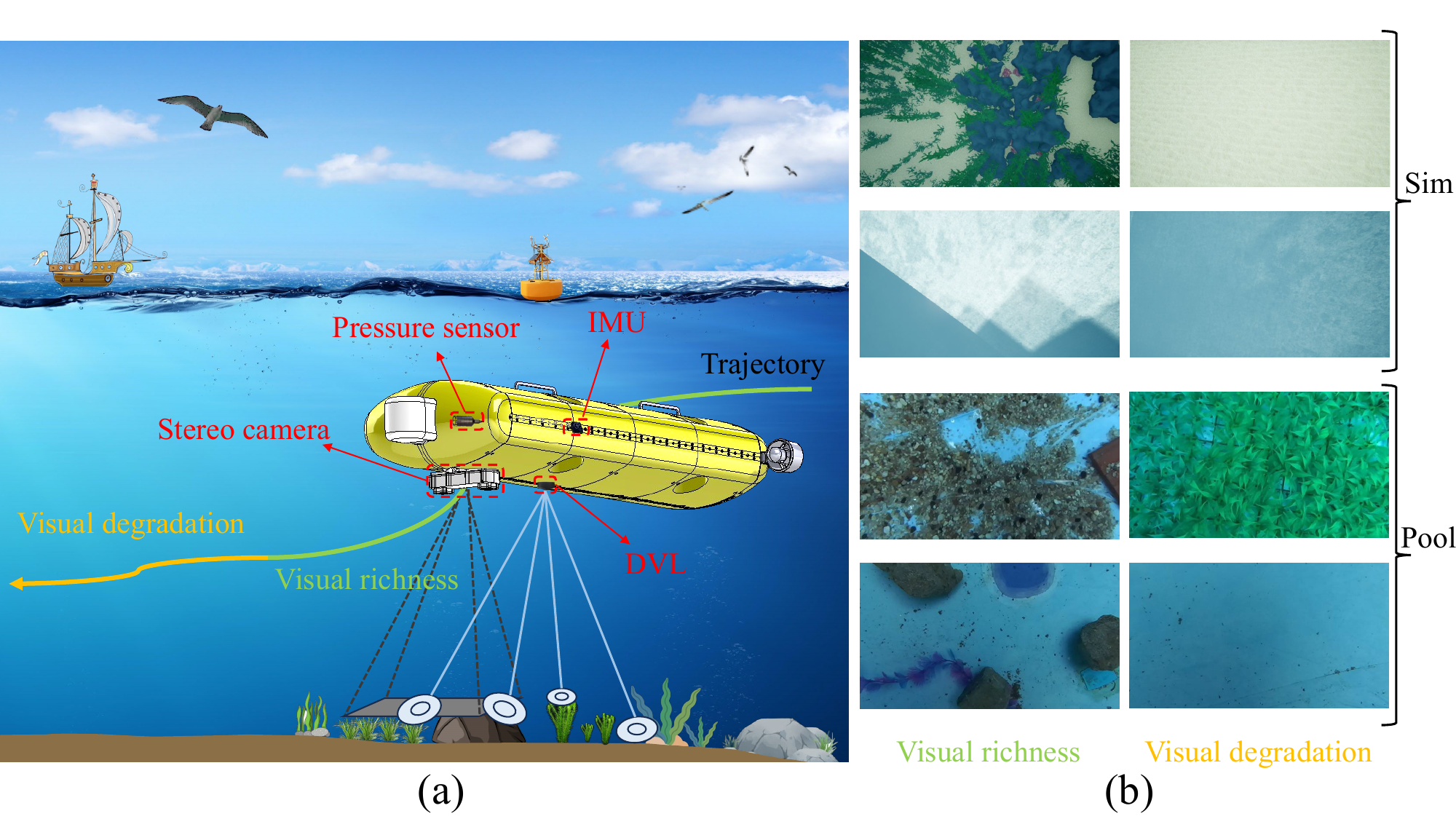}
	\caption{(a) Application scenarios for the proposed system and sensors related to positioning and perception. In visual richness regions, the system achieves stable pose estimation. When operating in visually degraded areas, the tight coupling of DVL, IMU and pressure sensor enables pose constraints, preventing abrupt trajectory drift or discontinuity. (b) Image examples from visually rich and visually degraded scenarios during underwater simulations and real-world pool experiments.}
	\label{fig1}
\end{figure}

The measurement of linear acceleration is coupled with both the gravitational direction and IMU bias, which further complicates the system \cite{xu2025aqua}. Therefore, to ensure accuracy and stability, an appropriate initialization process is crucial for effectively estimating the gravitational direction and IMU bias. This process typically relies on ideal visual conditions and sufficient motion excitation. However, achieving such conditions in underwater environments is exceedingly challenging. Furthermore, as shown in the visual degradation region of Fig. \ref{fig1}(b), the prolonged complete loss of structural and feature information in open water renders data from the IMU and pressure sensors insufficient to maintain stable operation of the system.

The Doppler velocity log (DVL) utilizes four-beam transmitters to measure real-time three-axis velocity data relative to the seabed and is commonly used for positioning and navigation in underwater environments. Unlike other sensors, DVL velocity data does not accumulate errors over time and operates across a wide range of velocities \cite{fan2024underwater}. DVL measurements remain effective even under conditions of visual loss. In underwater structured environments, Chavez et al. \cite{chavez2019adaptive} proposed a strategy that combines planar and point features by integrating ORB-SLAM2 \cite{mur2017orb}, DVL, and IMU within an extended Kalman filter (EKF) fusion system. Additionally, a Multi-State Constrained Kalman Filter was employed for state estimation by fusing visual imagery, DVL, IMU, and pressure sensor data \cite{zhao2023tightly}. However, in these fusion methods, measurements from multiple sensors are not simultaneously utilized during system state updates, and constraints cannot be established between sensors of different modalities.

Graph-based nonlinear optimization methods store information from motion and observation constraints in a graph consisting of nodes and edges. By optimizing these constraints, node estimates and uncertainties are obtained. A loosely coupled visual-acoustic bundle adjustment (BA) system is introduced, integrating the camera and acoustic odometer within a graph framework \cite{vargas2021robust}, \cite{xu2021underwater}. The loosely coupled strategy provides a simple structure and ease of implementation but fails to fully exploit the complementary advantages of the various sensor data. Huang et al. \cite{huang2024visual} fused DVL with visual inertial measurements. The gyroscope and DVL measurements were fused through an EKF and jointly optimized with the visual and inertial residuals. This tightly coupled strategy significantly enhanced the system's accuracy. Xu et al. \cite{xu2025aqua} conducted detailed investigations into multi-sensor external calibration and DVL transducer misalignment calibration, implementing real-time online operation via a rapid linear approximation procedure. The novel DVL calibration and modeling, combined with tightly coupled formulations, significantly improved the accuracy of the fusion system.

Despite existing efforts to enhance the accuracy of vision-based SLAM systems through sensor fusion, certain shortcomings persist. Previous research has primarily focused on the impact of underwater environments on visual sensors, with insufficient attention given to improving the visual tracking and optimization algorithms themselves. When enhancing positioning accuracy by integrating IMU and DVL, the information from the DVL is often underutilized, resulting in increased system redundancy. More critically, when confronted with visual degradation caused by extreme underwater environments, the attitude estimation of the VIO system rapidly drifts due to IMU noise. In this paper, we aim to mitigate drift by improving the state estimation during periods of visual degradation through the fusion of DVL, IMU, and pressure measurements.

In light of the aforementioned limitations, this paper proposes a graph-based underwater SLAM system that tightly integrates stereo camera, IMU, DVL, and pressure data. To address the issue of feature tracking loss underwater, a hybrid enhanced tracking method is introduced. The relative depth information acquired from the pressure sensor helps mitigate the IMU under-excitation problem. A velocity-bias-based DVL preintegration strategy is proposed. In particular, under conditions of visual degradation, tightly coupled constraints between the DVL, IMU, and pressure sensors facilitate state prediction, ensuring system stability. The main contributions of this article are in three aspects:
\begin{enumerate}[1)]
	\item A DVL preintegration method utilizing velocity bias is proposed to improve the operational efficiency and accuracy of DVL measurements. A hybrid tracking strategy, combining both indirect and direct methods, is employed during the front-end tracking process. Furthermore, the DVL's velocity measurements and position preintegration data are tightly integrated with IMU and pressure sensor data to mitigate localization drift and system failures during visual degradation.
	\item Within the local mapping thread, the visual hybrid residual, which includes both reprojection and photometric errors, is integrated with the inertial residual, DVL residual, and relative depth residual in a graph-based, tightly coupled optimization framework. These multi-source, heterogeneous constraints effectively improve the accuracy of pose estimation.
	\item Several different scenario datasets were gathered from both simulated and real-world underwater situations. The efficacy and accuracy of the suggested technique were validated using both qualitative and quantitative analysis. Notably, our technology displayed remarkable stability and robustness, especially in areas with visual degradation.
\end{enumerate}

The rest of the paper is organized as follows. The overview of our system is described in Section II. Section III illustrates the proposed viusal, inertial, acoustic and depth fusion method. Section IV provides the experiments and results. Finally, some conclusions are summarized in
Section V.

\section{SYSTEM OVERVIEW}

\begin{figure}[t]
	\centering
	\includegraphics[width=\linewidth]{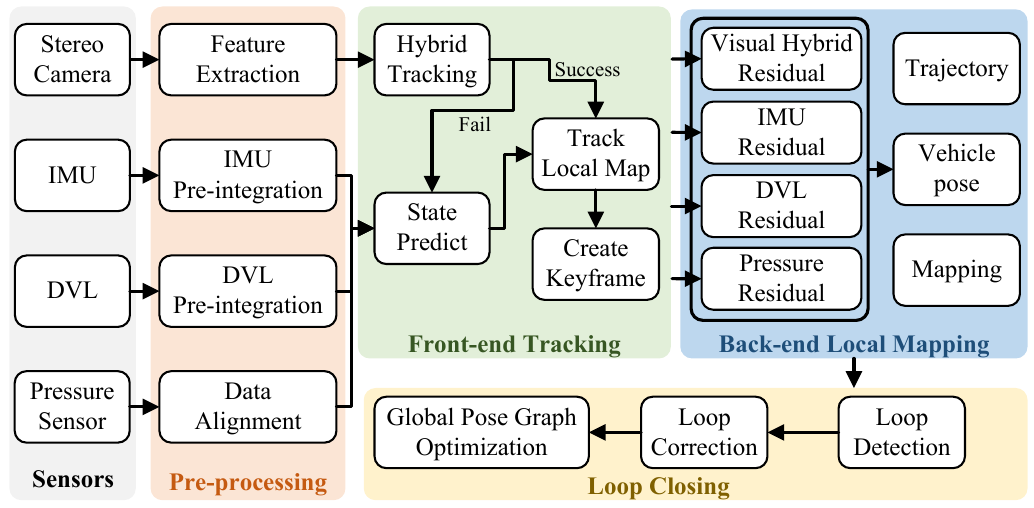}
	\caption{Multi-sensor fusion underwater SLAM framework.}
	\label{fig2}
\end{figure}

The overall architecture of the underwater SLAM system proposed herein, which integrates data from a stereo camera, an IMU, a DVL, and a pressure sensor, is depicted in Fig. \ref{fig2}. Data from the multiple sensors undergoes preliminary processing. The system extracts features from stereo images. The IMU and DVL measurements are pre-integrated, and the pressure data is processed. In the tracking thread, a hybrid tracking strategy that combines feature-based and direct methods is proposed. When visual tracking fails, pose estimation is carried out using IMU rotation data and DVL translation information, followed by local map tracking and optimization. In the local mapping thread, visual hybrid residuals are tightly coupled with IMU, DVL, and pressure residuals within a factor graph optimization framework. Finally, the loop-closing thread performs loop detection and global pose graph optimization. Inspired by ORB-SLAM3 \cite{campos2021orb}, these modules operate concurrently across multiple threads to ensure the system's operational rate and real-time performance.

Table \ref{tab1} shows the main notations used in this paper. Bold lower-case letters represent vectors and bold upper-case letters represent matrices. The transformation from camera frame to world frame is represented by a transform matrix $\mathbf{T}_\mathrm{WC} \in \mathrm{SE(3)}$, which can also be abbreviated as $\mathbf{T}_\mathrm{WC}=\left [ \mathbf{R}_\mathrm{WC},\mathbf{p}_\mathrm{WC} \right ] $, in which 3D rotation matrix is $\mathbf{R}_\mathrm{WC} \in \mathrm{SO(3)}$ and 3D translation is $\mathbf{p}_\mathrm{WC} \in \mathbb{R}^{3}$. The intensity image is denoted by $I:\Omega \mapsto \mathbb{R}$, where $\Omega\subset \mathbb{R}^{2} $ is the image domain. The camera projection function is represented by $ \pi \left ( \cdot  \right ) :\mathbb{R}^{3}\mapsto \Omega $, and the back-projection by $ \pi ^{-1} \left ( \cdot  \right ) : \Omega \times \mathbb{R} \mapsto \mathbb{R}^{3} $. $ \left ( \cdot  \right )^{\wedge}$ denotes a skew-symmetric matrix.

The state of frame $i$ is defined as:
\begin{align}
\mathrm{x}_{i}=\left \{\mathbf{R}_\mathrm{WI_\mathit{i}}, \mathbf{p}_\mathrm{WI_\mathit{i}}, \mathbf{v}_\mathrm{WI_\mathit{i}}, \mathbf{b}^{a}_{i},  \mathbf{b}^{g}_{i},  \mathbf{b}^{v}_{i} \right \} 
\label{equ1}
\end{align}
where $\left [ \mathbf{R}_\mathrm{WI_\mathit{i}}, \mathbf{p}_\mathrm{WI_\mathit{i}} \right ] $ and $\mathbf{v}_\mathrm{WI_\mathit{i}} \in \mathbb{R}^{3}$ denotes the IMU pose and velocity of frame $i$ relative to the world coordinate system. $\mathbf{b}^{a}_{i}$, $\mathbf{b}^{g}_{i}$ are the IMU accelerometer and gyroscope biases, respectively. $\mathbf{b}^{v}_{i}$ is the bias for DVL velocity measurement.

\begin{table}[t]
\caption{Glossary of notation.}
\label{tab1}
\begin{tabular}{l l}
\hline
Symbol & Meaning                                   \\ \hline
$\left ( \cdot \right )_{\mathrm{W}}$   & World frame         \\
$\left ( \cdot \right )_{\mathrm{I}}$   & IMU / body frame      \\
$\left ( \cdot \right )_{\mathrm{C}}$   & Camera frame        \\
$\left ( \cdot \right )_{\mathrm{D}}$   & DVL frame    \\
$\left ( \cdot \right )_{\mathrm{P}}$   & Pressure sensor frame    \\
$\left ( \cdot \right )_{\mathrm{W}_{i}}$   & The $i$th frame measured data in the world frame\\
$\mathbf{T}$           & Transform matrix \\
$\mathbf{R}$           & Rotation matrix                             \\
$\mathbf{p}$          & Translation vector                          \\
$\mathbf{v}$          & Velocity vector                             \\
$\mathrm{SE(3)}$      & The special Euclidean group              \\
$\mathrm{SO(3)}$      & The 3D special orthogonal group                                     \\ 
$\mathrm{Exp}$        & $\mathbb{R}^{3} \to \mathrm{SO(3)}$                           \\
$\mathrm{Log}$       & $\mathrm{SO(3)} \to \mathbb{R}^{3}$                           \\
\hline
\end{tabular}
\end{table}

\section{PROPOSED VISUAL-INERTIAL-ACOUSTIC-DEPTH FUSION}

\subsection{Data Preprocessing}

\textit{1) Visual Feature Extraction:} After capturing stereo images, ORB features are initially extracted from both images, followed by the matching of features between the left and right images. The depth of the feature points is then estimated through triangulation, based on the pixel position differences between the matched features, in preparation for subsequent map point optimization.

\textit{2) IMU Preintegration:} The data acquired from the IMU consists of discrete angular velocity measurements $\tilde{\mathbf{\omega}}_\mathrm{I}$ and accelerometer measurements $\tilde{\mathbf{a}}_\mathrm{I}$. $\mathbf{n}^{g}$ and $\mathbf{n}^{a}$ represent the Gaussian white noise measured by the gyroscope and accelerometer, $\mathbf{g}$ denotes the gravitational acceleration. $\Delta t$ is the sampling period of the IMU. Assuming the bias between two keyframes remains constant, the following relative motion increments are defined according to \cite{forster2016manifold}: 
\begin{align}
\Delta \mathbf{R}_\mathrm{I_\mathit{i} I_\mathit{j}} & \doteq \mathbf{R}_\mathrm{WI_\mathit{i}}^\mathrm{T} \mathbf{R}_\mathrm{WI_\mathit{j}} =  \prod\limits_{k=i}^{j-1} \operatorname{Exp}\left(\left(\tilde{\mathbf{\omega}}_\mathrm{I_\mathit{k}}-\mathbf{b}_{i}^{g}-\mathbf{n}_{k}^{g}\right) \Delta t\right)  \nonumber\\
\Delta \mathbf{v}_\mathrm{I_\mathit{i} I_\mathit{j}} &\doteq  \mathbf{R}_\mathrm{WI_\mathit{i}}^\mathrm{T}\left ( \mathbf{v}_\mathrm{WI_\mathit{j}}-\mathbf{v}_\mathrm{WI_\mathit{i}}-\mathbf{g} \Delta t_{ij} \right ) \nonumber\\ 
& =\sum\limits_{k=i}^{j-1} \Delta \mathbf{R}_\mathrm{I_\mathit{i} I_\mathit{k}}\left(\tilde{\mathbf{a}}_\mathrm{I_\mathit{k}}-\mathbf{b}_{i}^{a}-\mathbf{n}_{k}^{a}\right) \Delta t \nonumber\\
\Delta \mathbf{p}_\mathrm{I_\mathit{i} I_\mathit{j}} &\doteq  \mathbf{R}_\mathrm{WI_\mathit{i}}^\mathrm{T} \left ( \mathbf{p}_\mathrm{WI_\mathit{j}} -\mathbf{p}_\mathrm{WI_\mathit{i}} -\mathbf{v}_\mathrm{WI_\mathit{i}} \Delta t_{ij} -\frac{1}{2} \mathbf{g} \Delta t^{2}_{ij}  \right )    \nonumber\\
& =\sum\limits_{k=i}^{j-1} \left[ \Delta \mathbf{v}_\mathrm{I_\mathit{i} I_\mathit{k}} \Delta t+ \frac{1}{2} \Delta \mathbf{R}_\mathrm{I_\mathit{i}I_\mathit{k}}\left(\tilde{\mathbf{a}}_\mathrm{I_\mathit{k}}-\mathbf{b}_{i}^{a}-\mathbf{n}_{k}^{a}\right) \Delta t^{2}\right]
\label{equ2}
\end{align}

By separating the noise in (\ref{equ2}), the IMU preintegration from frame $i$ to $j$ can be obtained:
\begin{align}
\Delta \mathbf{\tilde{R}}_\mathrm{I_\mathit{i} I_\mathit{j}} &\doteq   \prod\limits_{k=i}^{j-1}\operatorname{Exp}\left(\left(\tilde{\mathbf{\omega}}_\mathrm{I_\mathit{k}}-\mathbf{b}_{i}^{g}\right) \Delta t\right) \nonumber\\
\Delta \mathbf{\tilde{v}}_\mathrm{I_\mathit{i} I_\mathit{j}} &\doteq  \sum\limits_{k=i}^{j-1}  \Delta \mathbf{\tilde{R}}_\mathrm{I_\mathit{i} I_\mathit{k}}\left(\tilde{\mathbf{a}}_\mathrm{I_\mathit{k}}-\mathbf{b}_{i}^{a}\right) \Delta t  \nonumber\\
\Delta \mathbf{\tilde{p}}_\mathrm{I_\mathit{i} I_\mathit{j}} &\doteq \sum\limits_{k=i}^{j-1} \left[\Delta \mathbf{\tilde{v}}_\mathrm{I_\mathit{i} I_\mathit{k}} \Delta t+ \frac{1}{2} \Delta \mathbf{\tilde{R}}_\mathrm{I_\mathit{i}I_\mathit{k}}\left(\tilde{\mathbf{a}}_\mathrm{I_\mathit{k}}-\mathbf{b}_{i}^{a}\right) \Delta t^{2}\right]
\label{equ3}
\end{align}

Equation (\ref{equ3}) represent the preintegration measurements of rotation, velocity, and translation, respectively. $\delta  \tilde{\phi}_\mathrm{I_\mathit{i} I_\mathit{j}} $ , $\delta  \mathbf{\tilde{v}}_\mathrm{I_\mathit{i} I_\mathit{j}} $ and $\delta  \mathbf{\tilde{p}}_\mathrm{I_\mathit{i} I_\mathit{j}}  $ correspond to their respective noise levels.

\textit{3) Pre-integrated DVL Measurements:} The DVL is equipped with four transducers oriented in different directions. The radial velocity of each transducer is determined based on the Doppler shift. Considering the spatial arrangement of the transducers, the DVL provides a set of three-dimensional data, representing velocity measurements along the three axes. This data is subsequently utilized by our system.

In the DVL coordinate system, the DVL velocity measurement associated with frame $i$ is denoted as $\mathbf{\tilde{v}}_\mathrm{D_\mathit{i}} \in \mathbb{R}^{3}$. Assuming a zero-mean Gaussian white noise $\mathbf{n}^{v}$. Inspired by the work on quadrupedal robot leg odometry \cite{wisth2022vilens}, we add a slowly varying bias term $\mathbf{b}^{v}_{i}$ to the velocity. Thus, the true DVL velocity is $\mathbf{v}_\mathrm{D_\mathit{i}}=\mathbf{\tilde{v}}_\mathrm{D_\mathit{i}}-\mathbf{b}^{v}_{i}-\mathbf{n}^{v}_{i}$. This bias incorporates drift caused by water bodies and inherent deviations. 

The external parameter transformation $\mathbf{T}_\mathrm{ID}=\left [ \mathbf{R}_\mathrm{ID},\mathbf{p}_\mathrm{ID} \right ]$ between the IMU and DVL represents the fixed relative pose between the two coordinate systems. By combining DVL velocity measurements with IMU orientation estimates, constraints can be applied to the translation estimates. Using the DVL velocity measurements between frames $i$ and $m$, the position can be iteratively computed in discrete time-domain form:
\begin{align}
\mathbf{p}_\mathrm{WD_\mathit{m}}=\mathbf{p}_\mathrm{WD_\mathit{i}}+\sum\limits_{k=i}^{m-1} \mathbf{R}_\mathrm{WD_\mathit{k}}\left ( \mathbf{\tilde{v}}_\mathrm{D_\mathit{k}}-\mathbf{b}^{v}_{i}-\mathbf{n}^{v}_{k}  \right ) \Delta t
\label{equ4}
\end{align}

From the (\ref{equ4}), it can be seen that the DVL position in frame $m$ is related to the pose of frame $i$. During optimization, changes in the pose of frame $i$ will lead to repeated integration of DVL transformations during iterations. Therefore, decouple the DVL translation estimation. Equation (\ref{equ4}) is multiplied by $\mathbf{R}^\mathrm{T}_\mathrm{WD_\mathit{i}}$ on both sides. From $\mathbf{T}_\mathrm{WD_\mathit{i}} =\mathbf{T}_\mathrm{WI_\mathit{i}}  \mathbf{T}_\mathrm{ID}$, we obtain $\mathbf{R}^\mathrm{T}_\mathrm{WD_\mathit{i}} =\mathbf{R}^\mathrm{T}_\mathrm{ID}  \mathbf{R}^\mathrm{T}_\mathrm{WI_\mathit{i}}$ and $\mathbf{p}_\mathrm{WD_\mathit{i}} =\mathbf{R}_\mathrm{WI_\mathit{i}}  \mathbf{p}_\mathrm{ID} + \mathbf{p}_\mathrm{WI_\mathit{i}}$. Furthermore, $\Delta \mathbf{R}_\mathrm{D_\mathit{i} D_\mathit{k}}=\mathbf{R}^\mathrm{T}_\mathrm{ID} \Delta \mathbf{R}_\mathrm{I_\mathit{i} I_\mathit{k}} \mathbf{R}_\mathrm{ID}$. Hence, we define the relative translation increment $\Delta \mathbf{p}_\mathrm{D_\mathit{i} D_\mathit{m}}$ of the DVL. Rewriting (\ref{equ4}) as:
\begin{align}
\Delta \mathbf{p}_\mathrm{D_\mathit{i} D_\mathit{m}} &\doteq \mathbf{R}^\mathrm{T}_\mathrm{WI_\mathit{i}} \left ( \left (  \mathbf{R}_\mathrm{WI_\mathit{m}} -\mathbf{R}_\mathrm{WI_\mathit{i}} \right ) \mathbf{p}_\mathrm{ID}  + \mathbf{p}_\mathrm{WI_\mathit{m}} - \mathbf{p}_\mathrm{WI_\mathit{i}}\right ) \nonumber\\
& = \sum\limits_{k=i}^{m-1} \Delta \mathbf{R}_\mathrm{I_\mathit{i} I_\mathit{k}} \mathbf{R}_\mathrm{ID}\left ( \mathbf{\tilde{v}}_\mathrm{D_\mathit{k}}-\mathbf{b}^{v}_{i}-\mathbf{n}^{v}_{k}  \right ) \Delta t
\label{equ5}
\end{align}

By substituting $\Delta  \mathbf{R}_\mathrm{I_\mathit{i} I_\mathit{k}}=\Delta \mathbf{\tilde{R}}_\mathrm{I_\mathit{i} I_\mathit{k}} \mathrm{Exp} ( -\delta  \tilde{\phi}_\mathrm{I_\mathit{i} I_\mathit{k}} )$ in (\ref{equ3}) to incorporate the pre-integrated rotation measurement, and the approximation $\mathrm{Exp}\left ( \phi   \right ) \simeq \mathbf{I}+\phi^{\wedge }$ and $\mathbf{a}^{\wedge }\mathbf{b}=-\mathbf{b}^{\wedge }\mathbf{a}$, $\forall \mathbf{a},\mathbf{b}\in \mathbb{R}^{3}$ . Separating the noise in (\ref{equ5}) becomes:
\begin{align}
\Delta \mathbf{p}_\mathrm{D_\mathit{i} D_\mathit{m}} \simeq  &\sum\limits_{k=i}^{m-1}  \Delta \mathbf{\tilde{R}}_\mathrm{I_\mathit{i} I_\mathit{k}} \mathbf{R}_\mathrm{ID}\left ( \mathbf{\tilde{v}}_\mathrm{D_\mathit{k}}-\mathbf{b}^{v}_{i}\right ) \Delta t  \nonumber\\
&- \sum\limits_{k=i}^{m-1}  [  -\Delta \mathbf{\tilde{R}}_\mathrm{I_\mathit{i} I_\mathit{k}} \left ( \mathbf{R}_\mathrm{ID}  \left ( \mathbf{\tilde{v}}_\mathrm{D_\mathit{k}}-\mathbf{b}^{v}_{i} \right )   \right )^{\wedge } \delta  \tilde{\phi}_\mathrm{I_\mathit{i} I_\mathit{k}} \Delta t  \nonumber\\
&+ \Delta \mathbf{\tilde{R}}_\mathrm{I_\mathit{i} I_\mathit{k}} \mathbf{R}_\mathrm{ID} \mathbf{n}^{v}_{k} \Delta t ]  \nonumber\\
\doteq &\Delta \mathbf{\tilde{p} }_\mathrm{D_\mathit{i} D_\mathit{m}} - \delta \mathbf{\tilde{p} }_\mathrm{D_\mathit{i} D_\mathit{m}}
\label{equ6}
\end{align}

Among these, we define the DVL translation preintegration measurement $\Delta \mathbf{\tilde{p} }_\mathrm{D_\mathit{i} D_\mathit{m}}$ and its noise $\delta \mathbf{\tilde{p} }_\mathrm{D_\mathit{i} D_\mathit{m}}$. Note that $\Delta \mathbf{\tilde{p} }_\mathrm{D_\mathit{i} D_\mathit{m}}$ still depends on the bias $\mathbf{b}^{g}_{i}$ and $\mathbf{b}^{v}_{i}$. When these are updated, it leads to recomputation. As done in \cite{forster2016manifold}, given a bias update $\mathbf{b} \gets \mathbf{\bar{b} } + \delta \mathbf{b} $, we can use a first-order expansion to update the preintegration.
\begin{align}
\Delta &\mathbf{\tilde{p} }_\mathrm{D_\mathit{i} D_\mathit{m}}(\mathbf{b}^{g}_{i},\mathbf{b}^{v}_{i}) \nonumber\\
= &\sum\limits_{k=i}^{m-1} \Delta \mathbf{\tilde{R}}_\mathrm{I_\mathit{i} I_\mathit{k}}(\mathbf{b}^{g}_{i})\mathbf{R}_\mathrm{ID}\left ( \mathbf{\tilde{v}}_\mathrm{D_\mathit{k}}-\mathbf{\bar{b}}^{v}_{i} - \delta \mathbf{b}^{v}_{i} \right ) \Delta t \nonumber\\
\simeq  &\sum\limits_{k=i}^{m-1} [\Delta \mathbf{\bar{R}}_\mathrm{I_\mathit{i} I_\mathit{k}}  (\mathbf{I} + ( \frac{\partial \Delta \mathbf{\bar{R}}_\mathrm{I_\mathit{i} I_\mathit{k}}}{\partial \mathbf{b}^{g}}  \delta \mathbf{b}^{g}_{i}  ) ^{\wedge } )   \nonumber\\
&\cdot  \mathbf{R}_\mathrm{ID} \left( \mathbf{\tilde{v}}_\mathrm{D_\mathit{k}}-\mathbf{\bar{b}}^{v}_{i} - \delta \mathbf{b}^{v}_{i}  \right) \Delta t ] \nonumber\\
\simeq &\sum\limits_{k=i}^{m-1} \Delta \mathbf{\bar{R}}_\mathrm{I_\mathit{i} I_\mathit{k}} \mathbf{R}_\mathrm{ID}   \left( \mathbf{\tilde{v}}_\mathrm{D_\mathit{k}}-\mathbf{\bar{b}}^{v}_{i} \right)\Delta t - \sum\limits_{k=i}^{m-1} \Delta \mathbf{\bar{R}}_\mathrm{I_\mathit{i} I_\mathit{k}} \mathbf{R}_\mathrm{ID}\Delta t \delta \mathbf{b}^{v}_{i} \nonumber\\
&- \sum\limits_{k=i}^{m-1} \Delta \mathbf{\bar{R}}_\mathrm{I_\mathit{i} I_\mathit{k}} (\mathbf{R}_\mathrm{ID} (\mathbf{\tilde{v}}_\mathrm{D_\mathit{k}}-\mathbf{\bar{b}}^{v}_{i}))^{\wedge } \Delta t \frac{\partial \Delta \mathbf{\bar{R}}_\mathrm{I_\mathit{i} I_\mathit{k}}}{\partial \mathbf{b}^{g}}  \delta \mathbf{b}^{g}_{i} \nonumber\\
= & \Delta \mathbf{\tilde{p} }_\mathrm{D_\mathit{i} D_\mathit{m}}(\mathbf{\bar{b}}^{g}_{i},\mathbf{\bar{b}}^{v}_{i}) + \frac{\partial \Delta \mathbf{\bar{p}}_\mathrm{D_\mathit{i} D_\mathit{m}}}{\partial \mathbf{b}^{v}}  \delta \mathbf{b}^{v}_{i} + \frac{\partial \Delta \mathbf{\bar{p}}_\mathrm{D_\mathit{i} D_\mathit{m}}}{\partial \mathbf{b}^{g}}  \delta \mathbf{b}^{g}_{i}
\label{equ7}
\end{align}
where, $\Delta \mathbf{\bar{R}}_\mathrm{I_\mathit{i} I_\mathit{k}}=\Delta \mathbf{\tilde{R}}_\mathrm{I_\mathit{i} I_\mathit{k}}(\mathbf{\bar{b}}^{g}_{i})$. $\frac{\partial \Delta \mathbf{\bar{p}}_\mathrm{D_\mathit{i} D_\mathit{m}}}{\partial \mathbf{b}^{v}}$ and $\frac{\partial \Delta \mathbf{\bar{p}}_\mathrm{D_\mathit{i} D_\mathit{m}}}{\partial \mathbf{b}^{g}}$ are the Jacobians used for the a posteriori bias update. When new measurements arrive, the Jacobians can be computed incrementally. Therefore, $\Delta \mathbf{\tilde{p} }_\mathrm{D_\mathit{i} D_\mathit{m}}(\mathbf{b}^{g}_{i},\mathbf{b}^{v}_{i})$ can be obtained by updating the preintegration translation measurement $\Delta \mathbf{\tilde{p} }_\mathrm{D_\mathit{i} D_\mathit{m}}(\mathbf{\bar{b}}^{g}_{i},\mathbf{\bar{b}}^{v}_{i})$, without requiring repeated integration.

\textit{4) Pressure Sensor Measurements:} The pressure measurement is derived from water pressure data, ultimately providing depth measurements relative to sea level. Since pressure represents a one-dimensional measurement along the gravitational direction, $\mathbf{s}3=[0,0,1]^\mathrm{T}$ is used to define the third dimension. Within the pressure sensor's coordinate system, the depth measurement $\mathrm{\tilde{p}}_\mathrm{p_\mathit{i}}$ at frame i is obtained. Based on the motion relationship between the pressure sensor and the IMU within this coordinate system, along with the IMU position estimate, the translational estimate for the pressure sensor can be determined:
\begin{align}
\mathbf{\hat{p}}_\mathrm{WP_\mathit{i}}=\mathbf{R}_\mathrm{WI_\mathit{i}} \mathbf{p}_\mathrm{IP}+\mathbf{p}_\mathrm{WI_\mathit{i}}
\label{equ8}
\end{align}

\subsection{Tracking Thread}

\begin{figure}[t]
	\centering
	\includegraphics[width=0.85\linewidth]{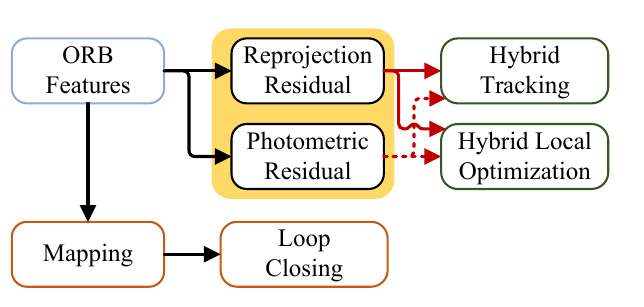}
	\caption{ORB features are employed in various forms within the proposed system.}
	\label{fig3}
\end{figure}

\textit{1) Visual Hybrid Tracking:} As illustrated in Fig. \ref{fig3}, a single ORB feature is extracted, and both the reprojection error from the feature-based method and the photometric error from the direct method are integrated into the feature tracking and local optimization modules.

Based on the pose $\mathbf{T}_\mathrm{WC_\mathit{i}}$ of the reference frame $i$ and the IMU preintegration information, the camera pose $\mathbf{T}_\mathrm{WC_\mathit{j}}$ of the current frame $j$ can be estimated using the uniform motion model. Using the estimated pose, 3D map points from the previous frame are projected onto the current frame, and matching feature points are searched for within a small neighborhood. The reprojection error between the current frame $j$ and the 3D map point $l$ projected onto frame $j$, can be defined as: $E_{jl}\doteq \mathbf{u}_{jl}-\pi \left ( \mathbf{T}_\mathrm{WC_\mathit{j}},l \right )$. $\mathbf{u}_{jl}$ denotes the pixel coordinates on frame $j$ of the feature point pair obtained through matching for landmark $l$.

Under the set of landmark points $L$ observed in the current frame $j$, the total reprojection error can be calculated as:
\begin{align}
E\left ( \mathbf{T}_\mathrm{WC_\mathit{j}} \right )_{repro} =\sum\limits_{l\in L}^{} \rho \left ( \left \| E_{jl}  \right \| ^{2} _\mathbf{P_\mathit{repro}} \right ) 
\label{equ9}
\end{align}
where $\mathbf{P_\mathit{repro}}$ denotes the covariance matrix of reprojection errors, derived from the measurement error and noise model. $\mathbf{T}_\mathrm{WC_\mathit{j}}$ represents the optimization objective for this step. The coarse pose plays a significant role in ensuring stable feature tracking.

Therefore, we perform a fast, direct sparse refinement of the coarse estimates provided by feature-based trackers. The results from the previous step provide an initial estimate that guides the direct method toward a region of rapid convergence.

For each feature point $\bm{p}$ on reference frame $i$ with known depth $d_{p}$, its predicted position $\bm{p}^{\prime}=\pi ( \mathbf{T}_\mathrm{C_\mathit{j} C_\mathit{i}},\pi^{-1}  ( \bm{p},d_{p} )  )$ on the current frame $j$. The photometric residual pattern is obtained from previous research \cite{engel2017direct}. $\mathcal{N}_{\bm{p}}$ represents the local pixel block. Based on the assumption of luminance constancy, the photometric residual is constructed as: $ E_{i{\bm{p}}j}\doteq \sum_{{\bm{p}} \in \mathcal{N}_{\bm{p}}}^{} w_{\bm{p}} (I_{j}(\bm{p}^{\prime})-I_{i}({\bm{p}}))$. $w_{\bm{p}}$ is a gradient-dependent weight that reduces the weight of pixels with high gradients. Therefore, for a set of feature points $\mathcal{P}_{i}$ within reference frame $i$, the total photometric residual is expressed as:
\begin{align}
E\left ( \mathbf{T}_\mathrm{WC_\mathit{j}} \right )_{photo} =\sum\limits_{\bm{p}\in L}^{} \rho \left ( \left \| E_{i{\bm{p}}j}  \right \| ^{2} _\mathbf{P_\mathit{photo}} \right )
\label{equ10}
\end{align}

The current frame pose $\mathbf{T}_\mathrm{WC_\mathit{j}}$ is refined again in this step.

\textit{2) State Predict:} When the vehicle enters a region of visual degradation, the VIO system loses the camera's ability to perceive the external environment. IMU propagation exhibits rapid drift due to IMU noise. To mitigate state drift during visual degradation, we propose fusing DVL and IMU data. Given the known pose $\mathbf{T}_\mathrm{WI_\mathit{i}}$ of the previous frame i, we estimate the pose $\mathbf{T}_\mathrm{WI_\mathit{j}}$ of the current frame $j$. The relative rotational increment $\mathbf{R}_\mathrm{WI_\mathit{j}}=\mathbf{R}_\mathrm{WI_\mathit{i}}\Delta \mathbf{R}_\mathrm{I_\mathit{i}I_\mathit{j}}$ is computed using (\ref{equ2}) for the IMU. For the translation component $\mathbf{p}_\mathrm{WI_\mathit{j}}$, we estimate it using the relative change derived from (\ref{equ5}) for the DVL.
\begin{align}
\mathbf{p}_\mathrm{WI_\mathit{j}}=\mathbf{R}_\mathrm{WI_\mathit{i}}\Delta \mathbf{p}_\mathrm{D_\mathit{i}D_\mathit{j}} + \mathbf{p}_\mathrm{WI_\mathit{i}}-(\mathbf{R}_\mathrm{WI_\mathit{j}}-\mathbf{R}_\mathrm{WI_\mathit{i}})\mathbf{p}_\mathrm{ID}
\label{equ11}
\end{align}

After successful frame-to-frame tracking, local map tracking is performed. In feature-rich regions, hybrid visual tracking operates stably, with camera, IMU, DVL, and pressure measurements being jointly optimized based on constraints between adjacent frames. In visually degraded regions, where visual tracking is lost, acoustic-inertial-depth optimization is applied to estimate the current frame's pose. This optimization leverages residuals from DVL, IMU, and pressure data, constrained by the relationships between the current frame and its reference keyframe.

\subsection{Local Mapping Thread}

\begin{figure}[t]
	\centering
	\includegraphics[width=1\linewidth]{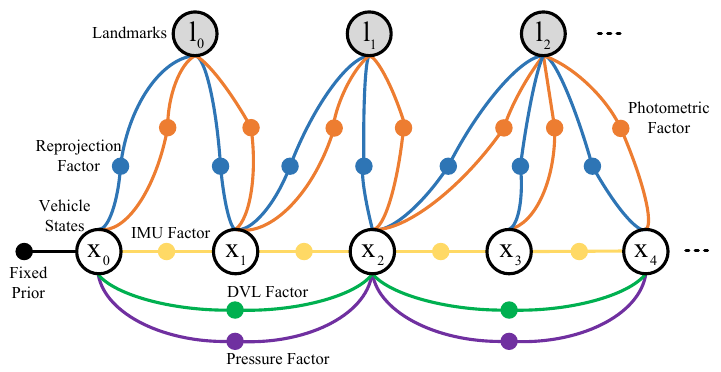}
	\caption{Factor graph structures in local optimization.}
	\label{fig4}
\end{figure}

Upon the arrival of a new keyframe, visual-inertial-acoustic-depth BA is performed within the local optimization window. As shown in Fig. \ref{fig4}, multi-sensor constraints are constructed based on the factor graph. Fixed prior factors are keyframes that are not within the local window but share a co-view relationship with the frame under optimization.

$Kf$ denotes a set of $k+1$ keyframes within a local window, while $L$ denotes a set of $l+1$ landmark points observed from these keyframes. Therefore, the variables to be optimized within the local window are written as:
\begin{align}
\chi =\left \{ \mathrm{x}_0,\mathrm{x}_1, \cdots  ,\mathrm{x}_k,\mathrm{l}_0,\mathrm{l}_1, \cdots, \mathrm{l}_l  \right \}
\label{equ12}
\end{align}
where $\mathrm{x}_k$ denotes the state of the $k$th keyframe, as shown in (\ref{equ1}). $\mathrm{l}_l$ represents the coordinates of a point in the 3-D map. Therefore, the local optimization problem can be expressed using the following overall cost function:
\begin{align}
\chi^{*} = &\underset{\chi}{\mathrm{argmin}}  ( \sum\limits_{i\in Kf} \sum\limits_{l\in L} \rho  ( \left \| E_{il}  \right \| _{\textbf{P}_{repro} }^{2} )  \nonumber\\
&+ \sum\limits_{i\in Kf} \sum\limits_{\bm{p}\in \mathcal{P}_{i}  } \sum\limits_{j\in obs ( \bm{p}  ) }\rho ( \left \| E_{i{\bm{p}}j}  \right \| _{\textbf{P}_{photo} }^{2} )  \nonumber\\
&+ \sum\limits_{i\in Kf}\left \| E_{i,i+1}  \right \|_{\textbf{P}_{imu}}^{2} + \sum\limits_{i,m\in Kf}\left \| E_{i,m}  \right \|_{\textbf{P}_{dvl}}^{2} \nonumber\\
&+ \sum\limits_{i,n\in Kf}\left \| E_{i,n}  \right \|_{\textbf{P}_{press}}^{2} )
\label{equ13}
\end{align}

Among these, $E_{il}$ and $E_{i{\bm{p}}j}$ are derived from (\ref{equ9}) and (\ref{equ10}), respectively, representing the reprojection residual and photometric residual. This approach introduces a dual-motion-constraint strategy, enabling the visual hybrid optimization model to achieve higher positioning accuracy with fewer feature points while also reducing computational cost.

Based on (\ref{equ3}), the IMU residual between consecutive keyframes $E_{i,i+1}\doteq [E_{\Delta \mathbf{R}_\mathrm{I_\mathit{i} I_\mathit{i\mathrm{+1}}}},E_{\Delta \mathbf{v}_\mathrm{I_\mathit{i} I_\mathit{i\mathrm{+1}}}},E_{\Delta \mathbf{p}_\mathrm{I_\mathit{i} I_\mathit{i\mathrm{+1}}}} ]$:
\begin{align}
E_{\Delta \mathbf{R}_\mathrm{I_\mathit{i} I_\mathit{i\mathrm{+1}}}}\doteq & \mathrm{Log}((\Delta \mathbf{\tilde{R}}_\mathrm{I_\mathit{i} I_\mathit{i\mathrm{+1}}}(\mathbf{b}^{g}_{i}))^\mathrm{T}\mathbf{R}^\mathrm{T}_\mathrm{W I_\mathit{i}}\mathbf{R}_\mathrm{W I_\mathit{i\mathrm{+1}}}) \nonumber\\
E_{\Delta \mathbf{v}_\mathrm{I_\mathit{i} I_\mathit{i\mathrm{+1}}}}\doteq &\mathbf{R}^\mathrm{T}_\mathrm{W I_\mathit{i}}(\mathbf{v}_\mathrm{W I_\mathit{i\mathrm{+1}}} - \mathbf{v}_\mathrm{W I_\mathit{i}} - \mathbf{g}\Delta t_{i,i\mathrm{+1}})  \nonumber\\
&-\Delta \mathbf{\tilde{v}}_\mathrm{I_\mathit{i} I_\mathit{i\mathrm{+1}}}(\mathbf{b}^{g}_{i},\mathbf{b}^{a}_{i}) \nonumber\\
E_{\Delta \mathbf{p}_\mathrm{I_\mathit{i} I_\mathit{i\mathrm{+1}}}}\doteq &\mathbf{R}^\mathrm{T}_\mathrm{W I_\mathit{i}}(\mathbf{p}_\mathrm{W I_\mathit{i\mathrm{+1}}} - \mathbf{p}_\mathrm{W I_\mathit{i}} -\mathbf{v}_\mathrm{W I_\mathit{i}}\Delta t_{ij} - \frac{1}{2} \mathbf{g}\Delta t^2_{i,i\mathrm{+1}})  \nonumber\\
&-\Delta \mathbf{\tilde{p}}_\mathrm{I_\mathit{i} I_\mathit{i\mathrm{+1}}}(\mathbf{b}^{g}_{i},\mathbf{b}^{a}_{i})
\label{equ14}
\end{align}

Based on the relationship between the DVL and IMU within the coordinate system, and utilizing the IMU velocity estimate, the DVL velocity estimate $\mathbf{\hat{v}}_\mathrm{D_\mathit{i}}=\mathbf{R}^\mathrm{T}_\mathrm{ID}(\mathbf{R}^\mathrm{T}_\mathrm{WI_\mathit{i}}  \mathbf{v}_\mathrm{WI_\mathit{i}} + (\tilde{\mathbf{\omega}}_\mathrm{I_\mathit{i}})^{\wedge }\mathbf{p}_\mathrm{ID})$ can be obtained \cite{fan2024underwater}. We utilize the relative velocity measurements and the translation preintegration from the DVL to constrain the relationships between keyframes. Based on (\ref{equ5}), an explicit expression for the DVL residual $E_{i,m}\doteq [E_{\Delta \mathbf{v}_\mathrm{D_\mathit{i} D_\mathit{m}}},E_{\Delta \mathbf{p}_\mathrm{D_\mathit{i} D_\mathit{m}}}]$ can be derived as:
\begin{align}
E_{\Delta \mathbf{v}_\mathrm{D_\mathit{i} D_\mathit{m}}}\doteq &(\mathbf{\hat{v}}_\mathrm{D_\mathit{m}}-\mathbf{\hat{v}}_\mathrm{D_\mathit{i}})-(\mathbf{\tilde{v}}_\mathrm{D_\mathit{m}}-\mathbf{\tilde{v}}_\mathrm{D_\mathit{i}}) \nonumber\\
E_{\Delta \mathbf{p}_\mathrm{D_\mathit{i} D_\mathit{m}}}\doteq &\mathbf{R}^\mathrm{T}_\mathrm{WI_\mathit{i}} \left ( \left (  \mathbf{R}_\mathrm{WI_\mathit{m}} -\mathbf{R}_\mathrm{WI_\mathit{i}} \right ) \mathbf{p}_\mathrm{ID}  + \mathbf{p}_\mathrm{WI_\mathit{m}} - \mathbf{p}_\mathrm{WI_\mathit{i}}\right )  \nonumber\\
&-\Delta \mathbf{\tilde{p} }_\mathrm{D_\mathit{i} D_\mathit{m}}(\mathbf{b}^{g}_{i},\mathbf{b}^{v}_{i})
\label{equ15}
\end{align}

According to (\ref{equ8}), the relative pressure residual between keyframes of the available depth \cite{ding2024underwater}:
\begin{align}
E_{i,n} &\doteq \mathbf{s}3(\mathbf{\hat{p}}_\mathrm{WP_\mathit{n}}-\mathbf{\hat{p}}_\mathrm{WP_\mathit{i}})-(\mathrm{\tilde{p}}_\mathrm{p_\mathit{n}}-\mathrm{\tilde{p}}_\mathrm{p_\mathit{i}}) \nonumber\\
&=\mathbf{s}3((\mathbf{R}_\mathrm{WI_\mathit{n}}-\mathbf{R}_\mathrm{WI_\mathit{i}})\mathbf{p}_\mathrm{IP}+\mathbf{p}_\mathrm{WI_\mathit{n}}-\mathbf{p}_\mathrm{WI_\mathit{i}})-(\mathrm{\tilde{p}}_\mathrm{p_\mathit{n}}-\mathrm{\tilde{p}}_\mathrm{p_\mathit{i}})
\label{equ16}
\end{align}

These residuals were used to formulate a least-squares problem based on Mahalanobis distance. Within the local optimization window, measurements from the camera, IMU, DVL, and pressure sensors were tightly coupled, fully exploiting the complementary strengths of each sensor to enhance the system's positioning accuracy and robustness.

\section{EXPERIMENTS AND RESULTS}

To thoroughly assess the feasibility of the proposed system, the Holo and HQPool datasets were collected from underwater simulators and real-world pool environments, respectively, both of which feature areas with significant visual degradation, as shown in Fig. \ref{fig1}(b). For comparison, two state-of-the-art stereo-inertial SLAM systems were selected: the feature-based ORB-SLAM3 (ORB3) \cite{campos2021orb} and the optical flow-based VINS-Fusion (VINS-F) \cite{qin2019general}. Additionally, our previous work \cite{ding2024underwater}, an underwater SLAM system integrating a monocular camera, IMU, and pressure sensor, was included in the comparison.

To more effectively evaluate the trajectories under analysis, the following preprocessing strategies were applied. During result storage, all trajectories were transformed into the IMU coordinate system. Since different systems have varying initialization times, a common timestamp was selected, and data prior to this time point were excluded. Simultaneously, the starting positions of all trajectories were translated to the origin. Additionally, to assess localization accuracy, the estimated trajectory was aligned with the ground truth at the starting point to minimize the mean squared error between the estimated trajectory and the ground truth \cite{umeyama2002least}. This ensures that all trajectories are consistently aligned and originate from the same reference point.

\subsection{Simulation Experiments}

\begin{table*}[t]
\caption{Performance comparison results on Holo dataset.}
\begin{tabular}{ccccccccc}
\hline
\multirow{2}{*}{\begin{tabular}[c]{@{}c@{}}Seq.\\ (Length)\end{tabular}} & \multicolumn{4}{c}{\begin{tabular}[c]{@{}c@{}}Translation error (in meter)\\ RMSE / STD\end{tabular}}         & \multicolumn{4}{c}{\begin{tabular}[c]{@{}c@{}}Rotation error (in degree)\\ RMSE / STD\end{tabular}}      \\ \cline{2-9} 
                                                                         & VINS-F        & ORB3          & \begin{tabular}[c]{@{}c@{}}Previous\\ Work\end{tabular} & Ours          & VINS-F         & ORB3          & \begin{tabular}[c]{@{}c@{}}Previous\\ Work\end{tabular} & Ours          \\ \hline
OW1 (108m)                                                               & 1.252 / 0.547 & 0.938 / 0.399 & 1.429 / 0.603                                           & 0.474 / 0.166 & 5.774 / 1.735  & 3.755 / 1.078 & 5.617 / 1.680                                           & 2.773 / 0.810 \\
OW2 (140m)                                                               & 4.360 / 1.822 & NaN / NaN     & NaN / NaN                                               & 1.175 / 0.505 & 11.576 / 4.386 & NaN / NaN     & NaN / NaN                                               & 3.662 / 1.371 \\
OW3 (246m)                                                               & 0.919 / 0.237 & 0.831 / 0.181 & 0.706 / 0.231                                           & 0.616 / 0.195 & 1.709 / 0.511  & 1.632 / 0.500 & 1.054 / 0.304                                           & 0.679 / 0.229 \\
OW4 (171m)                                                               & 0.426 / 0.124 & 0.445 / 0.157 & 0.523 / 0.172                                           & 0.352 / 0.106 & 1.047 / 0.301  & 1.311 / 0.327 & 1.637 / 0.382                                           & 0.953 / 0.262 \\
Dam1 (268m)                                                              & 5.767 / 2.051 & 2.595 / 0.921 & 3.503 / 1.328                                           & 1.499 / 0.565 & 29.708 / 7.624 & 7.119 / 2.079 & 12.861 / 6.031                                          & 5.817 / 1.567 \\
Dam2 (113m)                                                              & 0.856 / 0.327 & 0.990 / 0.398 & 1.041 / 0.590                                           & 0.562 / 0.234 & 2.920 / 1.050  & 3.310 / 1.38  & 3.481 / 1.384                                           & 2.228 / 0.908 \\
Dam3 (195m)                                                              & NaN / NaN     & NaN / NaN     & NaN / NaN                                               & 1.828 / 0.709 & NaN / NaN      & NaN / NaN     & NaN / NaN                                               & 6.034 / 2.989 \\
Dam4 (176m)                                                              & 1.470 / 0.466 & 1.173 / 0.409 & 1.218 / 0.381                                           & 0.687 / 0.194 & 7.03 / 2.13    & 5.232 / 1.860 & 4.901 / 1.437                                           & 2.582 / 0.729 \\ \hline
Avg.*                                                                    & 1.782 / 0.625 & 1.162 / 0.411 & 1.402 / 0.551                                           & 0.698 / 0.243 & 8.03 / 2.22    & 3.726 / 1.205 & 4.925 / 1.870                                           & 2.505 / 0.750 \\ \hline
\multicolumn{9}{l}{$*$The successful sequences for all systems have been calculated.}                           
\end{tabular}
\label{tab2}
\end{table*}

\begin{figure}[t]
	\centering
	\includegraphics[width=1\linewidth]{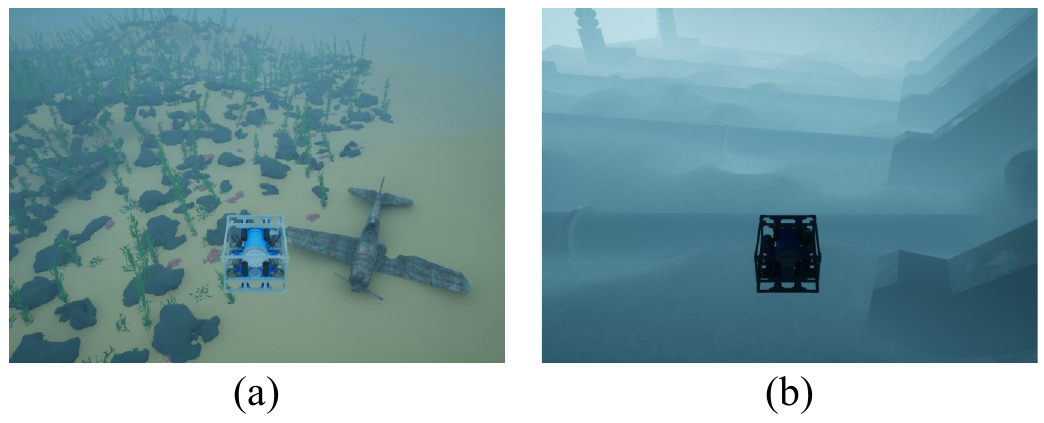}
	\caption{Underwater simulation environment. (a) Open-Water scenes. (b) Dam scenes.}
	\label{fig5}
\end{figure}

The Holo simulation datasets were acquired from the HoloOcean \cite{potokar2024holoocean} underwater simulator, as shown in Fig. \ref{fig5}. The Open-Water (OW) scenario includes aquatic plants, rocks, and a crashed aircraft, providing rich visual features. In contrast, the Dam scenario consists primarily of a few pipes and dam structures, with sparse textures and significant visual degradation. Multi-sensor data were collected from the ROV by the simulator. The stereo camera operated at a resolution of 640 $\times$ 360 pixels with a frame rate of 15 Hz, while the IMU functioned at 100 Hz. Both the DVL and pressure sensor recorded data at 10 Hz.

\begin{figure}[t]
	\centering
	\includegraphics[width=1\linewidth]{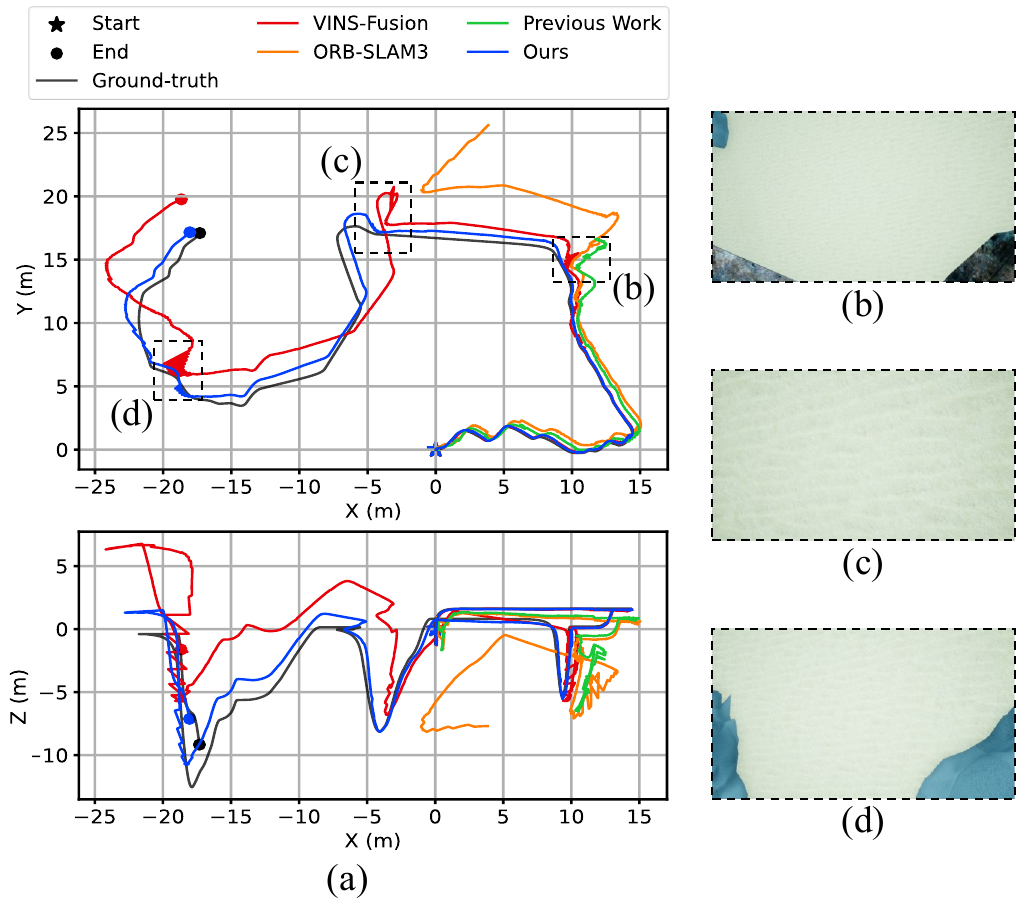}
	\caption{Trajectory results on the OW2 sequence. (a) shows the trajectory comparison between the X-Y axis and X-Z axis. (b), (c), and (d) display challenging images within the visual degradation region.}
	\label{fig6}
\end{figure}

\begin{figure}[t]
	\centering
	\includegraphics[width=1\linewidth]{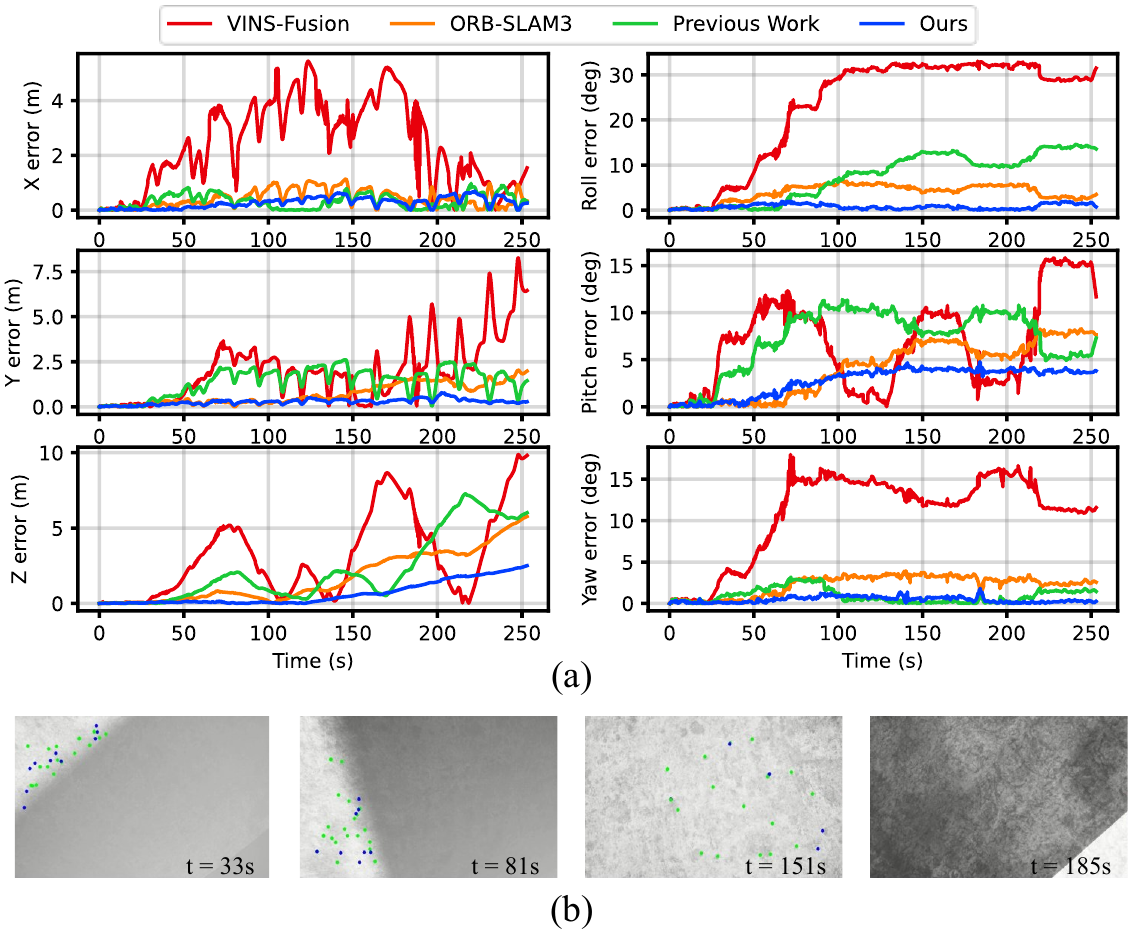}
	\caption{Comparison of error results on the Dam1 sequence. (a) Translation errors in X-Y-Z and rotation errors in roll-pitch-yaw over time. (b) Example of feature tracking by our system in visually degraded regions.}
	\label{fig7}
\end{figure}

\textit{1) Quantitative Evaluation:} In the simulation environment, precise ground truth are available, allowing for the reporting of both absolute translation and rotation errors. To evaluate system performance in noisy environments, the root-mean-square errors (RMSE) and standard deviations (STD) of these errors are provided in Table \ref{tab2}. System failures were observed for VINS-F, ORB3, and our previous work. Prolonged visual degradation occurred in sequences OW2 and Dam3, as shown in Figs. \ref{fig6}(b)-(d) and Fig. \ref{fig7}(b). Thanks to DVL-IMU state prediction and acoustic-inertial-depth optimization, our proposed system remained operational throughout all sequences.

Overall, in sequences where both systems succeeded, ORB3 outperformed VINS-F due to its efficient data association and optimization techniques. In comparison, the previous work did not demonstrate superior performance relative to either ORB3 or VINS-F. The stereo-inertial system, benefiting from the advantages of stereo matching, significantly outperforms the monocular-inertial system. The contribution of pressure sensor measurements to the monocular-inertial system remains limited. Regarding RMSE, our system reduces translation errors by 60.8\% and 39.9\% compared to VINS-F and ORB3, respectively, and reduces rotation errors by 68.8\% and 32.7\%.

These results clearly demonstrate that the proposed strategy offers substantial improvements in both translation and rotation accuracy. This can be attributed to the proposed SLAM method, which tightly integrates visual, inertial, acoustic, and depth data in both the front-end and back-end processing stages.

\textit{2) Qualitative Analysis:} As shown in Fig. \ref{fig6}(a), both ORB3 and previous methods employ feature-based approaches but fail to complete the entire trajectory. This limitation is primarily due to the insufficient availability of tracking features in visually degraded regions. Both systems exhibit significant drift at the location indicated in Fig. \ref{fig6}(b). While ORB3 acquires more features using stereo images, it fails when encountering another featureless region, as shown in Fig. \ref{fig6}(c). The optical flow-based VINS-F demonstrates greater robustness in textureless scenes; however, prolonged periods of featureless tracking cause pose estimation to become highly unstable, as illustrated in Figs. \ref{fig6}(c) and \ref{fig6}(d).

Constrained by pressure and DVL measurements, our approach also limits Z-axis drift. The proposed system demonstrates greater robustness than other strategies when encountering visually degraded scenarios. This is primarily due to the hybrid tracking method enhancing feature reliability, while state predictions from DVL and IMU improve tracking stability.

The error comparison in Fig. \ref{fig7}(a) shows that our system achieves low error in both 3D translation and 3-axis rotation. Notably, around 30 s into operation, the VINS-F system exhibits significant drift, primarily due to the ROV's large-scale motion, which degrades the performance of the optical flow tracking strategy. Fig. \ref{fig7}(b) illustrates the feature tracking performance of the proposed system during visually degraded periods. The tracked features are sparsely and unevenly distributed, which is highly detrimental to pose estimation. In particular, at 185 s, feature tracking is temporarily lost. Despite this, our system maintains strong stability and accuracy, further validating the effectiveness of the proposed multi-sensor tightly coupled optimization.

\begin{figure}[t]
	\centering
	\includegraphics[width=1\linewidth]{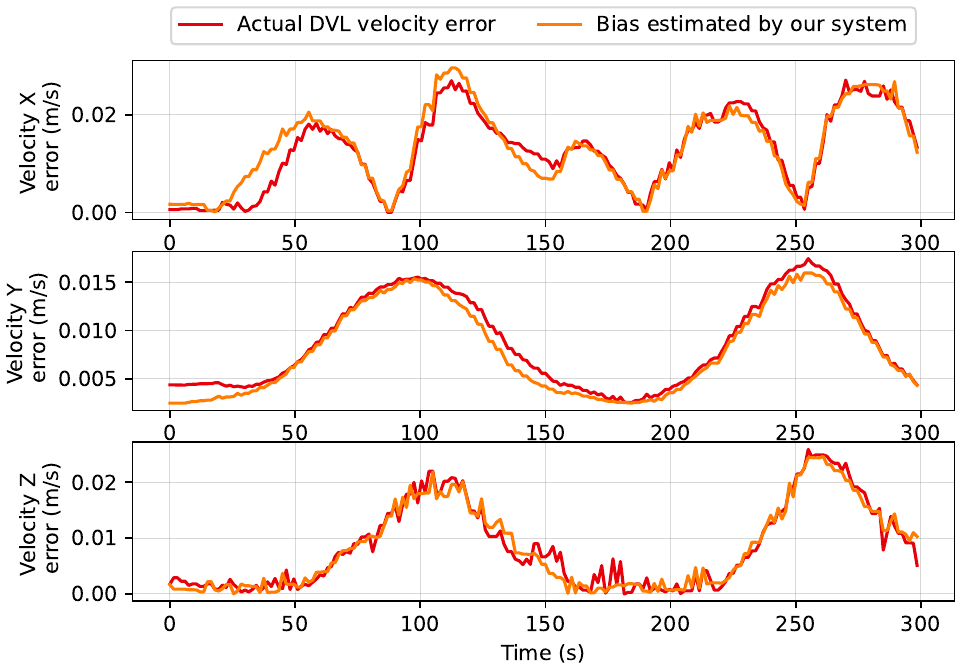}
	\caption{Comparison between actual DVL velocity errors and bias estimated by our system during the OW4 experiment.}
	\label{fig8}
\end{figure}

\textit{3) DVL Velocity Bias Estimation Accuracy:} In Fig. \ref{fig8}, we compare the actual velocity error and the estimated velocity error along the X, Y, and Z axes for the OW4 sequence. These errors represent drift in DVL velocity measurements caused by moving objects or water flow. Velocity bias estimation is employed to correct these drift errors. The red line represents the difference between the actual vehicle velocity and the DVL-measured velocity, while the orange line shows the estimated error, as calculated by the proposed method, modeled as the velocity bias term $\mathbf{b}^{v}$ within the DVL factor. As depicted, a high correlation is observed between these two signal sets, demonstrating the effectiveness of the DVL velocity bias estimation and validating the high-precision application of DVL measurements within the system.

\subsection{Real-World Experiments}

\begin{figure}[t]
	\centering
	\includegraphics[width=1\linewidth]{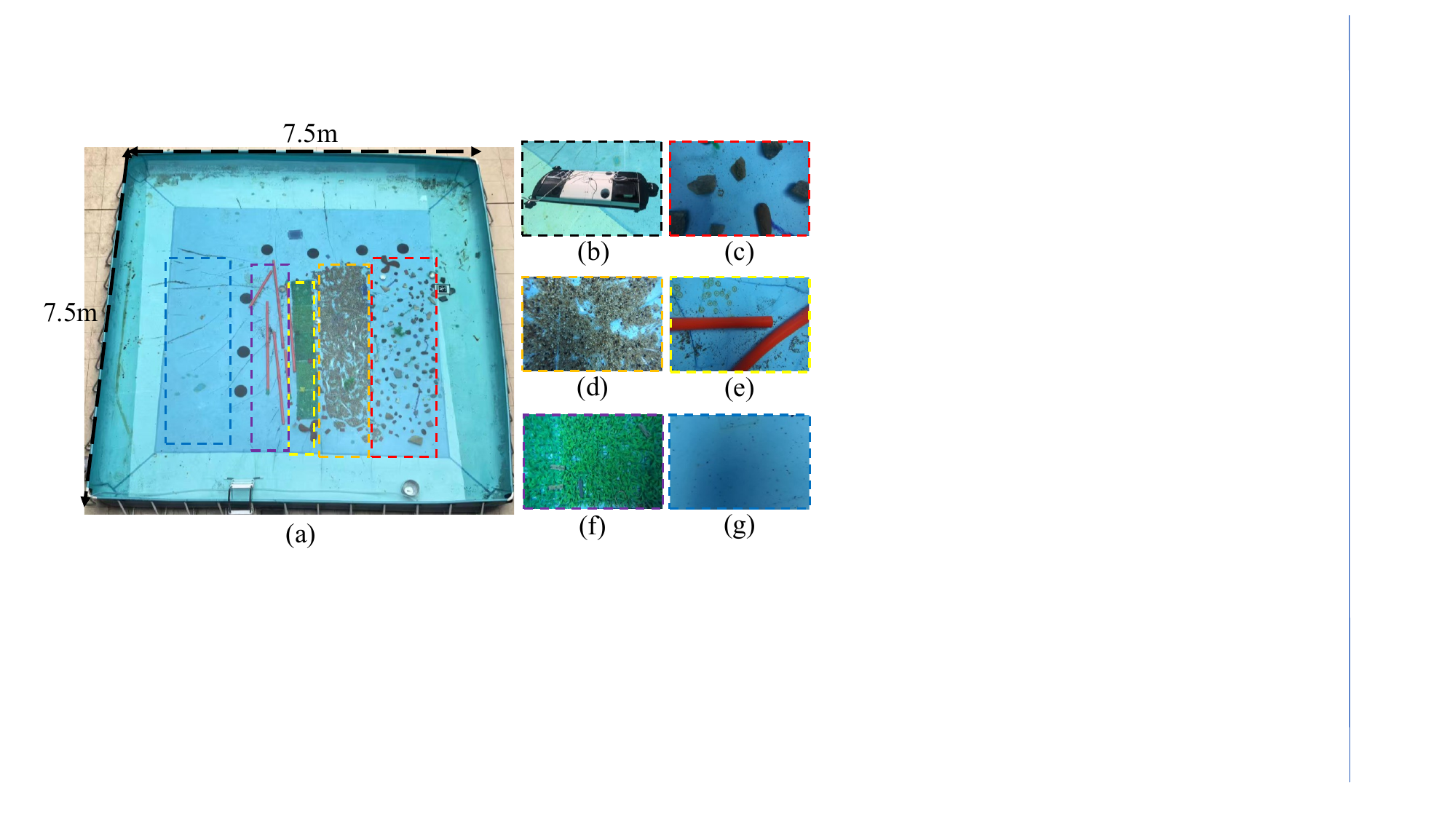}
	\caption{Experimental scenario of the self-collected HQPool dataset. (a) Aerial view of the experimental pool. (b) Self-developed underwater vehicle. (c)-(g) Examples of self-built underwater scenes.}
	\label{fig9}
\end{figure}

The HQPool dataset was collected in a real-world underwater environment. The experimental setup is depicted in Fig. \ref{fig9}(a). Various objects were placed in the pool to simulate different seabed scenarios, including rocky bottoms, sandy bottoms, grassy bottoms, and areas with subsea pipelines. Additionally, most regions remained unobstructed to simulate a muddy seabed. Figs. \ref{fig9}(c) to (g) illustrate that the rocky and pipeline regions exhibit rich visual features, while the sandy and grassy substrates display highly repetitive textures. In contrast, the muddy substrate contains extremely sparse features, which pose significant challenges for the tested systems.

Fig. \ref{fig9}(b) shows the self-developed underwater vehicle. During the equipment installation, the assembly strictly adhered to the mechanical design drawings, thereby establishing the conversion relationships between the sensors. The actual motion trajectory of the vehicle was captured by the global vision system installed above the pool. The stereo camera mounted on the vehicle operates at a frame rate of 15 Hz and a resolution of 672 $\times$ 376 pixels. The IMU operates at 100 Hz, the DVL at 9 Hz, and the pressure sensor at 10 Hz.

\begin{table}[t]
\caption{Comparison of translation errors on the HQPool dataset (RMSE / STD in meter).}
\begin{tabular}{ccccc}
\hline
\begin{tabular}[c]{@{}c@{}}Seq.\\ (Length)\end{tabular} & VINS-F                            & ORB3                              & \begin{tabular}[c]{@{}c@{}}Previous \\ Work\end{tabular} & Ours                              \\ \hline
HQ1 (29m)                                             & 0.299/0.098                     & 0.265/0.075                     & 0.473/0.137                                            & 0.222/0.069                     \\
HQ2 (50m)                                            & 0.477/0.219                     & 0.406/0.192                     & 0.540/0.216                                            & 0.245/0.101                     \\
HQ3 (83m)                                             & 0.583/0.247                     & 0.478/0.203                     & 0.527/0.198                                            & 0.365/0.135                     \\
HQ4 (55m)                                             & 1.183/0.358                     & 1.341/0.587                     & NaN/NaN                                                & 0.534/0.173                     \\
HQ5 (132m)                                            & 3.711/1.186                     & NaN/NaN                         & NaN/NaN                                                & 0.807/0.293                     \\ \hline
Avg.*    & \multicolumn{1}{l}{0.453/0.188} & \multicolumn{1}{l}{0.383/0.157} & \multicolumn{1}{l}{0.513/0.184}                        & \multicolumn{1}{l}{0.277/0.102} \\ \hline
\multicolumn{5}{l}{$*$The successful sequences for all systems have been calculated.}                          
\end{tabular}
\label{tab3}
\end{table}

\begin{figure*}[t]
	\centering
	\includegraphics[width=1\linewidth]{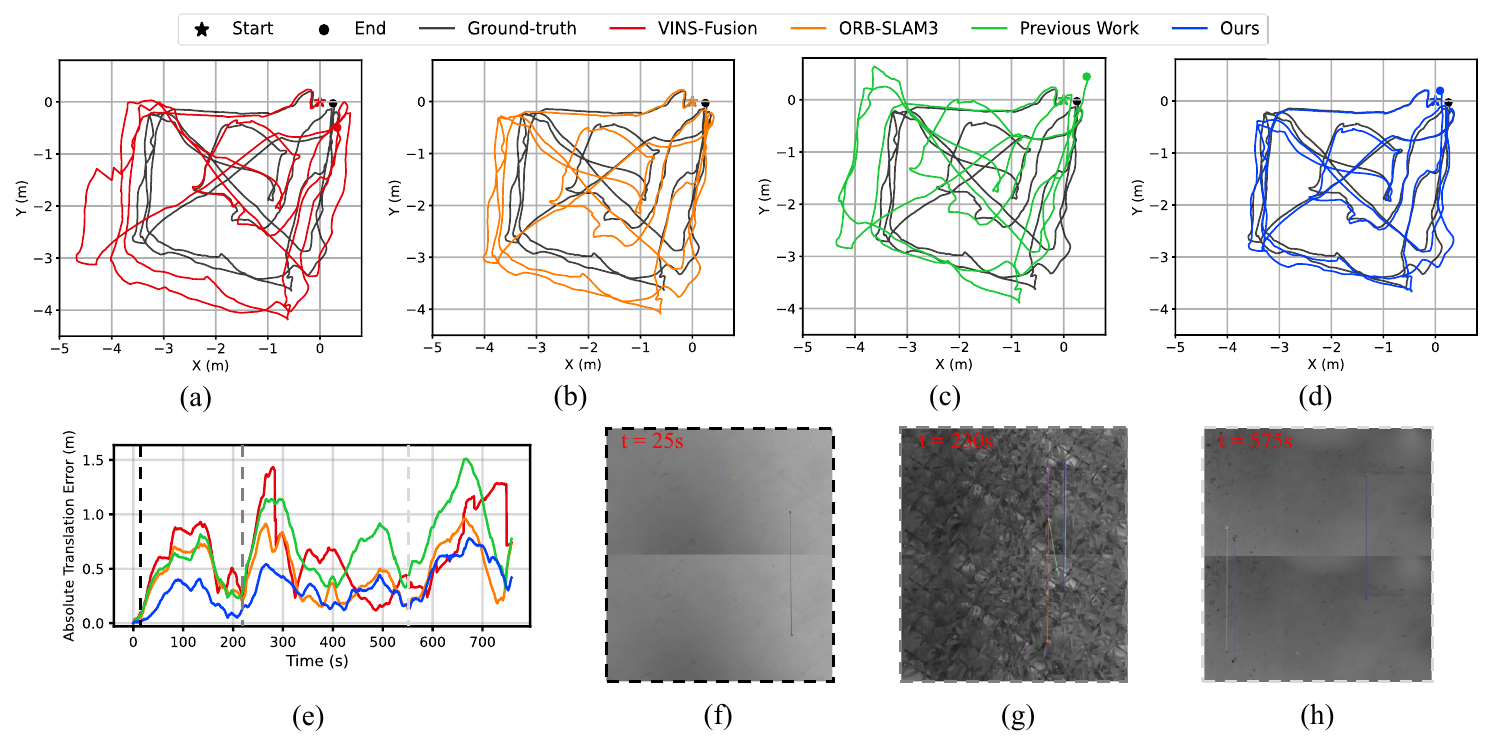}
	\caption{Experimental results on HQ2 sequence. (a) to (d) show the X-Y trajectory comparisons of the tested systems. (e) plots the absolute trajectory error over time. (f), (g), and (h) demonstrate feature matching examples within the visual degradation areas.}
	\label{fig10}
\end{figure*}

\textit{1) Quantitative Evaluation:} Table \ref{tab3} presents the translation errors of the tested systems on the HQPool dataset. In the pool experiments, the UUV motion is relatively slow and constrained by the water depth, resulting in minimal vertical movement. Our previous work was unable to achieve accurate scale estimation, leading to significant translation errors. Stereo images, through feature matching, provide advantages in scale estimation and can generate more stable map points. On the visually rich HQ1 sequence, both VINS-F and ORB3 achieved satisfactory positioning performance. However, on the HQ4 and HQ5 sequences, these systems exhibited significant drift and even failure when encountering large-scale visual degradation areas, as shown in Figs. \ref{fig10}(f) and \ref{fig10}(g). Our system, which fuses visual-inertial data with DVL and pressure measurements, demonstrates superior positioning accuracy and robustness.

On the three successful sequences, our system achieved an average RMSE reduction of 38.9\% and 27.7\% compared to VINS-F and ORB3, respectively, while the average STD decreased by 45.7\% and 35\%. A comparison of the mean translation errors shows that our method reduces the STD more significantly than the RMSE. This is primarily because the STD is more sensitive to fluctuations in error. In regions of short-duration visual degradation, the VINS-F and ORB3 visual-inertial systems rely more heavily on IMU estimates, leading to abrupt drifts. However, as visual information becomes more abundant, the rate of error growth decreases. The proposed DVL and pressure constraints help mitigate rapid IMU drifts, thereby enhancing system stability.

\textit{2) Qualitative Analysis:} The X-Y trajectory comparisons of the tested systems in sequence HQ2 are presented in Figs. \ref{fig10}(a) to (d) for qualitative analysis. Near the starting point, all systems successfully closed loops, and the trajectories were globally optimized and corrected. However, due to error accumulation, ORB3 exhibited limited adjustment throughout the entire trajectory, with significant drift persisting far from the origin. In contrast, our method demonstrates superior global trajectory tracking performance, benefiting from the joint constraints of visual hybrid residuals, IMU residuals, DVL residuals, and pressure sensor residuals.

During the three time intervals marked in Fig. \ref{fig10}(e), all systems exhibited significant error growth. This is attributed to the visual degradation areas, which contain extremely sparse image features, as shown in Figs. \ref{fig10}(f) and \ref{fig10}(h). In regions with repetitive textures, as depicted in Fig. \ref{fig10}(g), feature matching becomes difficult, leading to misalignments. Under these conditions, visual-inertial systems struggle to provide reliable pose estimates. In contrast, the tight coupling of DVL, IMU, and pressure sensors in our approach mitigates the rapid error growth caused by visual degradation.

\section{CONCLUSIONS}

To enhance the robustness and accuracy of visual-inertial systems in visually degraded environments, this paper proposes a novel underwater SLAM system that integrates stereo camera, IMU, DVL, and pressure measurements within a graph optimization framework. A novel velocity-bias-based DVL preintegration strategy is also introduced. The proposed velocity bias facilitates stable tracking of true values, effectively leveraging DVL measurements for pose estimation and enhancing the accuracy of data utilization.

At the system frontend, the proposed visual hybrid tracking strategy and the DVL-IMU-pressure state prediction and optimization method significantly enhance tracking stability. Particularly during prolonged periods of visual degradation, the system maintains stable operation, demonstrating notable robustness. Visual hybrid residuals, inertial residuals, DVL residuals, and pressure residuals are tightly integrated within a factor graph, thereby improving the system's optimization performance. In real-world underwater experiments, the proposed system achieves positioning accuracy that is 38.9\% and 27.7\% higher compared to VINS-Fusion and ORB-SLAM3, respectively.

In future work, DVL data alignment will be further investigated to enhance measurement effectiveness. Additionally, the sparse point clouds obtained from DVL hold significant importance for localization and mapping.


\bibliographystyle{Bibliography/IEEEtran}

\bibliography{Bibliography/IEEEabrv,Bibliography/BIB_xx-tran-xxxx}\ 

\begin{thebibliography}{10}
\providecommand{\url}[1]{#1}
\csname url@samestyle\endcsname
\providecommand{\newblock}{\relax}
\providecommand{\bibinfo}[2]{#2}
\providecommand{\BIBentrySTDinterwordspacing}{\spaceskip=0pt\relax}
\providecommand{\BIBentryALTinterwordstretchfactor}{4}
\providecommand{\BIBentryALTinterwordspacing}{\spaceskip=\fontdimen2\font plus
\BIBentryALTinterwordstretchfactor\fontdimen3\font minus
  \fontdimen4\font\relax}
\providecommand{\BIBforeignlanguage}[2]{{%
\expandafter\ifx\csname l@#1\endcsname\relax
\typeout{** WARNING: IEEEtran.bst: No hyphenation pattern has been}%
\typeout{** loaded for the language `#1'. Using the pattern for}%
\typeout{** the default language instead.}%
\else
\language=\csname l@#1\endcsname
\fi
#2}}
\providecommand{\BIBdecl}{\relax}
\BIBdecl

\bibitem{bai2025sio}
J.~Bai, D.~Zhu, M.~Chen, and C.~Luo, ``Sio-uv: Rapid and robust sonar intertial
  odometry for underwater vehicles,'' \emph{IEEE Trans. Ind. Electron.},
  \href{http://dx.doi.org/10.1109/TIE.2025.3561817}{DOI
  10.1109/TIE.2025.3561817}, pp. 1--11, May. 2025.

\bibitem{ma2020efficient}
T.~Ma, Y.~Li, Y.~Zhao, Y.~Jiang, Q.~Zhang, and P.~Ant{\'o}nio~M, ``Efficient
  bathymetric slam with invalid loop closure identification,'' \emph{IEEE/ASME
  Trans. Mechatron.}, vol.~26, no.~5, pp. 2570--2580, Oct. 2020.

\bibitem{xie2025neurss}
Y.~Xie, J.~Zhang, N.~Bore, and J.~Folkesson, ``Neurss: Enhancing auv
  localization and bathymetric mapping with neural rendering for sidescan
  slam,'' \emph{IEEE J. Ocean. Eng.}, vol.~50, no.~3, pp. 1596--1605, Jul.
  2025.

\bibitem{rahman2022svin2}
S.~Rahman, A.~Quattrini~Li, and I.~Rekleitis, ``Svin2: A multi-sensor
  fusion-based underwater slam system,'' \emph{Int. J. Robot. Res.}, vol.~41,
  no. 11-12, pp. 1022--1042, Jul. 2022.

\bibitem{ding2024robust}
S.~Ding, T.~Zhang, M.~Lei, H.~Chai, and F.~Jia, ``Robust visual-based
  localization and mapping for underwater vehicles: A survey,'' \emph{Ocean
  Eng.}, vol. 312, no. 11-12, p. 119274, Nov. 2024.

\bibitem{ou2024hybrid}
Y.~Ou, J.~Fan, C.~Zhou, P.~Zhang \emph{et~al.}, ``Hybrid-vins: Underwater
  tightly coupled hybrid visual inertial dense slam for auv,'' \emph{IEEE
  Trans. Ind. Electron.}, vol.~72, no.~3, pp. 2821--2831, Mar. 2024.

\bibitem{miao2021univio}
R.~Miao, J.~Qian, Y.~Song, R.~Ying, and P.~Liu, ``Univio: Unified direct and
  feature-based underwater stereo visual-inertial odometry,'' \emph{IEEE Trans.
  Instrum. Meas.}, vol.~71, pp. 1--14, Dec. 2021.

\bibitem{wang2023robust}
Y.~Wang, D.~Gu, X.~Ma, J.~Wang, and H.~Wang, ``Robust real-time auv
  self-localization based on stereo vision-inertia,'' \emph{IEEE Trans. Veh.
  Technol.}, vol.~72, no.~6, pp. 7160--7170, Jun. 2023.

\bibitem{hu2022tightly}
C.~Hu, S.~Zhu, Y.~Liang, and W.~Song, ``Tightly-coupled
  visual-inertial-pressure fusion using forward and backward imu
  preintegration,'' \emph{IEEE Robot. Autom. Lett.}, vol.~7, no.~3, pp.
  6790--6797, Jul. 2022.

\bibitem{ding2023rd}
S.~Ding, T.~Ma, Y.~Li, S.~Xu, and Z.~Yang, ``Rd-vio: Relative-depth-aided
  visual-inertial odometry for autonomous underwater vehicles,'' \emph{Appl.
  Ocean Res.}, vol. 134, p. 103532, May. 2023.

\bibitem{xu2025aqua}
S.~Xu, K.~Zhang, and S.~Wang, ``Aqua-slam: Tightly-coupled underwater
  acoustic-visual-inertial slam with sensor calibration,'' \emph{IEEE Trans.
  Robot.}, vol.~41, pp. 2785--2803, Mar. 2025.

\bibitem{fan2024underwater}
J.~Fan, X.~Liu, Y.~Ou, P.~Zhang, C.~Zhou, and Z.~Hou, ``Underwater robot
  self-localization method using tightly coupled events, images, inertial, and
  acoustic fusion,'' \emph{IEEE Trans. Ind. Electron.}, vol.~72, no.~5, pp.
  5126--5135, Oct. 2024.

\bibitem{chavez2019adaptive}
A.~G. Chavez, Q.~Xu, C.~A. Mueller, S.~Schwertfeger, and A.~Birk, ``Adaptive
  navigation scheme for optimal deep-sea localization using multimodal
  perception cues,'' in \emph{IEEE/RSJ Int. Conf. Intell. Robot. Syst. (IROS)},
  pp. 7211--7218, Nov. 2019.

\bibitem{mur2017orb}
R.~Mur-Artal and J.~D. Tard{\'o}s, ``Orb-slam2: An open-source slam system for
  monocular, stereo, and rgb-d cameras,'' \emph{IEEE Trans. Ind. Electron.},
  vol.~33, no.~5, pp. 1255--1262, Jun. 2017.

\bibitem{zhao2023tightly}
L.~Zhao, M.~Zhou, and B.~Loose, ``Tightly-coupled visual-dvl-inertial odometry
  for robot-based ice-water boundary exploration,'' in \emph{IEEE/RSJ Int.
  Conf. Intell. Robot. Syst. (IROS)}, pp. 7127--7134, Oct. 2023.

\bibitem{vargas2021robust}
E.~Vargas, R.~Scona, J.~S. Willners, T.~Luczynski, Y.~Cao, S.~Wang, and Y.~R.
  Petillot, ``Robust underwater visual slam fusing acoustic sensing,'' in
  \emph{IEEE Int. Conf. Rob. Autom. (ICRA)}, pp. 2140--2146, May. 2021.

\bibitem{xu2021underwater}
S.~Xu, T.~Luczynski, J.~S. Willners, Z.~Hong, K.~Zhang, Y.~R. Petillot, and
  S.~Wang, ``Underwater visual acoustic slam with extrinsic calibration,'' in
  \emph{IEEE/RSJ Int. Conf. Intell. Robot. Syst. (IROS)}, pp. 7647--7652, Sep.
  2021.

\bibitem{huang2024visual}
Y.~Huang, P.~Li, S.~Ma, S.~Yan, M.~Tan, J.~Yu, and Z.~Wu,
  ``Visual-inertial-acoustic sensor fusion for accurate autonomous localization
  of underwater vehicles,'' \emph{IEEE Trans. Cybern.}, vol.~55, no.~2, pp.
  880--896, Nov. 2024.

\bibitem{campos2021orb}
C.~Campos, R.~Elvira, J.~J.~G. Rodr{\'\i}guez, J.~M. Montiel, and J.~D.
  Tard{\'o}s, ``Orb-slam3: An accurate open-source library for visual,
  visual--inertial, and multimap slam,'' \emph{IEEE Trans. Robot.}, vol.~37,
  no.~6, pp. 1874--1890, May. 2021.

\bibitem{forster2016manifold}
C.~Forster, L.~Carlone, F.~Dellaert, and D.~Scaramuzza, ``On-manifold
  preintegration for real-time visual--inertial odometry,'' \emph{IEEE Trans.
  Robot.}, vol.~33, no.~1, pp. 1--21, Aug. 2016.

\bibitem{wisth2022vilens}
D.~Wisth, M.~Camurri, and M.~Fallon, ``Vilens: Visual, inertial, lidar, and leg
  odometry for all-terrain legged robots,'' \emph{IEEE Trans. Robot.}, vol.~39,
  no.~1, pp. 309--326, Aug. 2022.

\bibitem{engel2017direct}
J.~Engel, V.~Koltun, and D.~Cremers, ``Direct sparse odometry,'' \emph{IEEE
  Trans. Pattern Anal. Mach. Intell.}, vol.~40, no.~3, pp. 611--625, Apr. 2017.

\bibitem{ding2024underwater}
S.~Ding, T.~Zhang, Y.~Li, S.~Xu, and M.~Lei, ``Underwater multi-sensor fusion
  localization with visual-inertial-depth using hybrid residuals and efficient
  loop closing,'' \emph{Measurement}, vol. 238, no.~3, p. 115245, Oct. 2024.

\bibitem{qin2019general}
T.~Qin, S.~Cao, J.~Pan, and S.~Shen, ``A general optimization-based framework
  for global pose estimation with multiple sensors,'' \emph{arXiv:1901.03642},
  2019.

\bibitem{umeyama2002least}
S.~Umeyama, ``Least-squares estimation of transformation parameters between two
  point patterns,'' \emph{IEEE Trans. Pattern Anal. Mach. Intell.}, vol.~13,
  no.~4, pp. 376--380, Apr. 2002.

\bibitem{potokar2024holoocean}
E.~Potokar, K.~Lay, K.~Norman, D.~Benham, S.~Ashford, R.~Peirce, T.~B. Neilsen,
  M.~Kaess, and J.~G. Mangelson, ``Holoocean: A full-featured marine robotics
  simulator for perception and autonomy,'' \emph{IEEE J. Ocean. Eng.}, vol.~49,
  no.~4, pp. 1322--1336, Aug. 2024.

\end{thebibliography}


\end{document}